\documentclass[final,1p,times,num]{elsarticle}

\usepackage[numbers]{natbib}
\usepackage{booktabs}
\usepackage{hyperref}
\usepackage{color}

\usepackage[ruled,linesnumbered]{algorithm2e}
\usepackage{amsmath}
\usepackage{bm}
\usepackage{seqsplit}
\usepackage{graphicx}
\usepackage{caption}
\usepackage{amsmath}

\def\tsc#1{\csdef{#1}{\textsc{\lowercase{#1}}\xspace}}
\tsc{WGM}
\tsc{QE}
\tsc{EP}
\tsc{PMS}
\tsc{BEC}
\tsc{DE}
\journal{  }

\begin{document}
\let\WriteBookmarks\relax
\def\floatpagepagefraction{1}
\def\textpagefraction{.001}

\begin{frontmatter}



\title{Back to Fundamentals: Low-Level Visual Features Guided Progressive Token Pruning} 


\author[1]{Yuanbing Ouyang}
\author[1]{Yizhuo Liang}
\author[1]{Qingpeng Li}
\author[2]{Xinfei Guo}
\author[1]{Yiming Luo}
\author[3]{Di Wu}
\author[1]{Hao Wang}
\author[4]{Yushan Pan}
\ead{Corresponding authors at Xi'an Jiaotong-Liverpool Univeristy, Email: yushan.pan@xjtlu.edu.cn}
\affiliation[1]{organization={Xidian University},
            addressline={No. 266, Xinglong Section, Xifeng Road}, 
            city={Xi'An},
            postcode={710126  }, 
            state={Shannxi},
            country={China}}

\affiliation[2]{organization={University of Michigan – Shanghai Jiao Tong University Joint Institute, Shanghai Jiao 
Tong University },
            addressline={800 Dong Chuan Road}, 
            city={Shanghai},
            postcode={200240  },
            country={China}}

\affiliation[3]{organization={Norwegian University of Science and Technology},
            addressline={Larsgaardsvegen 2}, 
            city={Aalesund},
            postcode={6009},
            country={Norway}}

\affiliation[4]{organization={Xi'an Jiaotong-Liverpool University},
            addressline={111 Ren’ai Road
Suzhou Industrial Park}, 
            city={Suzhou},
            postcode={215123}, 
            state={Jiangsu},
            country={China}}
\begin{abstract}
Vision Transformers (ViTs) excel in semantic segmentation but demand significant computation, posing challenges for deployment on resource-constrained devices. Existing token pruning methods often overlook fundamental visual data characteristics. This study introduces `\textbf{LVTP}', a progressive token pruning framework guided by multi-scale Tsallis entropy and low-level visual features with twice clustering. It integrates high-level semantics and basic visual attributes for precise segmentation. A novel dynamic scoring mechanism using multi-scale Tsallis entropy weighting overcomes limitations of traditional single-parameter entropy. The framework also incorporates low-level feature analysis to preserve critical edge information while optimizing computational cost. As a plug-and-play module, it requires no architectural changes or additional training. Evaluations across multiple datasets show 20\%-45\% computational reductions with negligible performance loss, outperforming existing methods in balancing cost and accuracy, especially in complex edge regions.
\end{abstract}


\end{frontmatter}



\section{Introduction}
\label{sec1}


Transformer-based large language models (LLMs) like GPT-4 and SAM are revolutionizing AI \cite{a2}\cite{a3}\cite{a4}. In computer vision, Vision Transformer (ViT) \cite{a47}\cite{a46} enables cross-modal feature fusion via global attention \cite{a191}, surpassing CNNs on benchmarks like CIFAR-10 \cite{a165}\cite{a24}. ViT- and UNet-based models excel in semantic segmentation \cite{a192}, but their high computational cost poses challenges.

Semantic segmentation, crucial for autonomous driving and medical imaging \cite{a97}, is shifting to edge computing. However, ViT’s fine-grained pixel processing leads to exponential computational overhead \cite{a29}, making energy-efficient deployment on mobile platforms like drones difficult \cite{b}\cite{b11}\cite{b19}.

Optimization strategies focus on two main techniques:

\begin{enumerate}
    \item \textbf{Quantization} compresses models by mapping high-precision parameters to lower precision \cite{b68}, but struggles with numerical sensitivity, multi-scale feature preservation, and hardware compatibility \cite{b21}.
    \item \textbf{Structured pruning} removes non-essential layers to improve efficiency \cite{b50}\cite{b}, yet risks disrupting critical feature pathways, especially in long-range dependency tasks.
\end{enumerate}

A promising alternative, token pruning, dynamically removes redundant tokens in ViT, leveraging self-attention’s ability to process variable-length sequences \cite{18}. Over 60\% of tokens in natural images contribute little to predictions, such as uniform sky regions in autonomous driving \cite{17}. Unlike CNN pruning, which maintains fixed tensor dimensions, ViT pruning enables adaptive computational graph optimization.

Most token pruning studies focus on classification and detection, prioritizing high-level semantic feature retention \cite{19}. However, semantic segmentation requires not only object recognition but also precise boundary delineation and pixel-wise classification \cite{20}\cite{21}. Classic models like UNet \cite{cao2022swin} enhance segmentation via skip connections, while Segformer \cite{21} integrates multi-scale features for efficiency. These approaches improve accuracy but increase computational costs.

Existing token pruning methods often overlook fine-grained texture details and spatial context, which are essential for segmentation. Chen’s experiments show that pruning solely based on high-level features significantly degrades segmentation accuracy \cite{30}. Addressing this limitation is critical for effective model compression in segmentation tasks.

To address these challenges, we propose LVTP (Low-Level Visual Features Guided Progressive Token Pruning), a novel approach for semantic segmentation. Unlike image classification, segmentation demands fine-grained spatial understanding. LVTP clusters tokens based on multi-scale entropy, initially retaining those with high semantic information. A secondary clustering step, guided by low-level image features, further refines the pruning process.

Our key contributions:

\begin{itemize}
    \item We introduce a \textbf{dynamic scoring mechanism} using multi-scale Tsallis entropy for task-oriented token importance assessment. A simple unsupervised clustering method evaluates semantic complexity, guiding pruning more effectively.
    \item We propose a \textbf{progressive pruning framework}, where initial pruning is entropy-weighted, and final pruning is refined via low-order feature bootstrapping. This layer-wise strategy balances efficiency and performance. To our knowledge, LVTP is the first to integrate low-level features into token pruning.
    \item LVTP requires \textbf{no additional training} or fine-tuning, making it a plug-and-play solution. Experiments show a \textbf{20\% reduction in GFLOPS with no performance loss} and only \textbf{3\% accuracy loss at 45\% pruning}, demonstrating its potential for real-world deployment.

\end{itemize}

The paper is structured as follows: Section 2 reviews related work. Section 3 details the methodology. Section 4 presents experiments, followed by a discussion in Section 5. Finally, Section 6 concludes the paper.

\section{Related Work}
\subsection{ ViT in Semantic Segmentation}
Typical image segmentation tasks, such as instance segmentation \cite{hafiz2020survey}\cite{bolya2019yolact} and semantic segmentation \cite{hatamizadeh2021swin}\cite{lu2023content}, assign semantic labels to each pixel, enabling networks to extract rich semantic information for deep visual understanding and precise image analysis. Vision Transformers (ViTs) have gained significant attention in semantic segmentation, leveraging self-attention mechanisms to effectively capture long-range dependencies. This enhances their ability to model global context and adapt to complex scenes, which is crucial for fine-grained segmentation \cite{20,23,24}.

With the rise of visual LLMs, semantic segmentation has advanced significantly \cite{25,27}. ViT-based foundation models, trained on large-scale datasets, exhibit strong feature extraction capabilities and generalization. As a result, researchers are increasingly fine-tuning these models for specific tasks, shifting semantic segmentation from general-purpose applications to task-oriented approaches \cite{26}. This transition necessitates more adaptive compression strategies.

\subsection{Efficient ViT}
Existing research primarily focuses on model quantization and pruning techniques. Quantization methods have advanced rapidly, with two primary approaches: quantization-aware training (QAT) and post-training quantization (PTQ). QAT optimizes quantization strategies during model retraining to improve compression performance \cite{zhao2023post}. Classic methods such as Q-BERTQ \cite{D57} and HAWQ \cite{D82} have achieved remarkable compression results. However, retraining large language models (LLMs) for QAT requires significant computational resources and time, limiting its practical deployment.  

Although PTQ eliminates the need for retraining, it has its own limitations. For instance, Liu et al. \cite{D21} proposed a mixed-precision quantization scheme that integrates ranking loss and nuclear norm-based methods. Lin et al. \cite{D22} introduced the Power-of-Two Factor (PTF) and Log-Int-Softmax (LIS) methods, while Ding et al. \cite{D42} developed the APQ-ViT framework. Despite their innovations, these methods often require additional training or involve complex quantization matching processes, making deployment challenging.  

Pruning techniques represent another key approach to model compression, divided into unstructured \cite{D144,D145} and structured pruning \cite{D148,D149,D152}. Unstructured pruning removes individual weights, while structured pruning eliminates entire model components, preserving performance more effectively but demanding significant computational resources to determine optimal pruning paths \cite{28}. Additionally, model importance and redundancy vary across different tasks and scenarios, resulting in poor adaptability and limiting the ability to optimize models effectively based on task-specific requirements.

\subsection{Token Pruning}
Significant progress has been made in token pruning for classification tasks, encompassing various approaches such as attention scoring \cite{C1,C2}, neural network- or reinforcement learning-assisted methods \cite{C3,C4,C5}, and rule-based techniques \cite{C6,C7,C8}. These strategies provide strong support for efficient sparse computation.

For ViT-based semantic segmentation, several studies have contributed notable advancements. DynamicViT \cite{29}, for example, employs a lightweight prediction module to estimate token importance and utilizes an attention mask strategy for pruning. VLTP \cite{30} introduces a pruning decoder guided by a multimodal large language model (MLLM) to enhance token pruning effectiveness. However, integrating MLLM increases computational complexity, making lightweight inference on edge devices challenging.

Despite these advancements, ViT compression for computationally intensive tasks still faces limitations. Many pruning methods require retraining or fine-tuning, restricting their adaptability across different architectures and scenarios. Furthermore, the predominant reliance on high-level semantics in pruning overlooks the critical role of low-level image features in delineating object boundaries and segmenting small objects. Developing a generalized token pruning method that incorporates low-level features while balancing computational cost and performance remains a key research challenge, essential for advancing semantic segmentation on edge devices.

\begin{figure*}[h!]
    \centering
    \includegraphics[width=\textwidth]{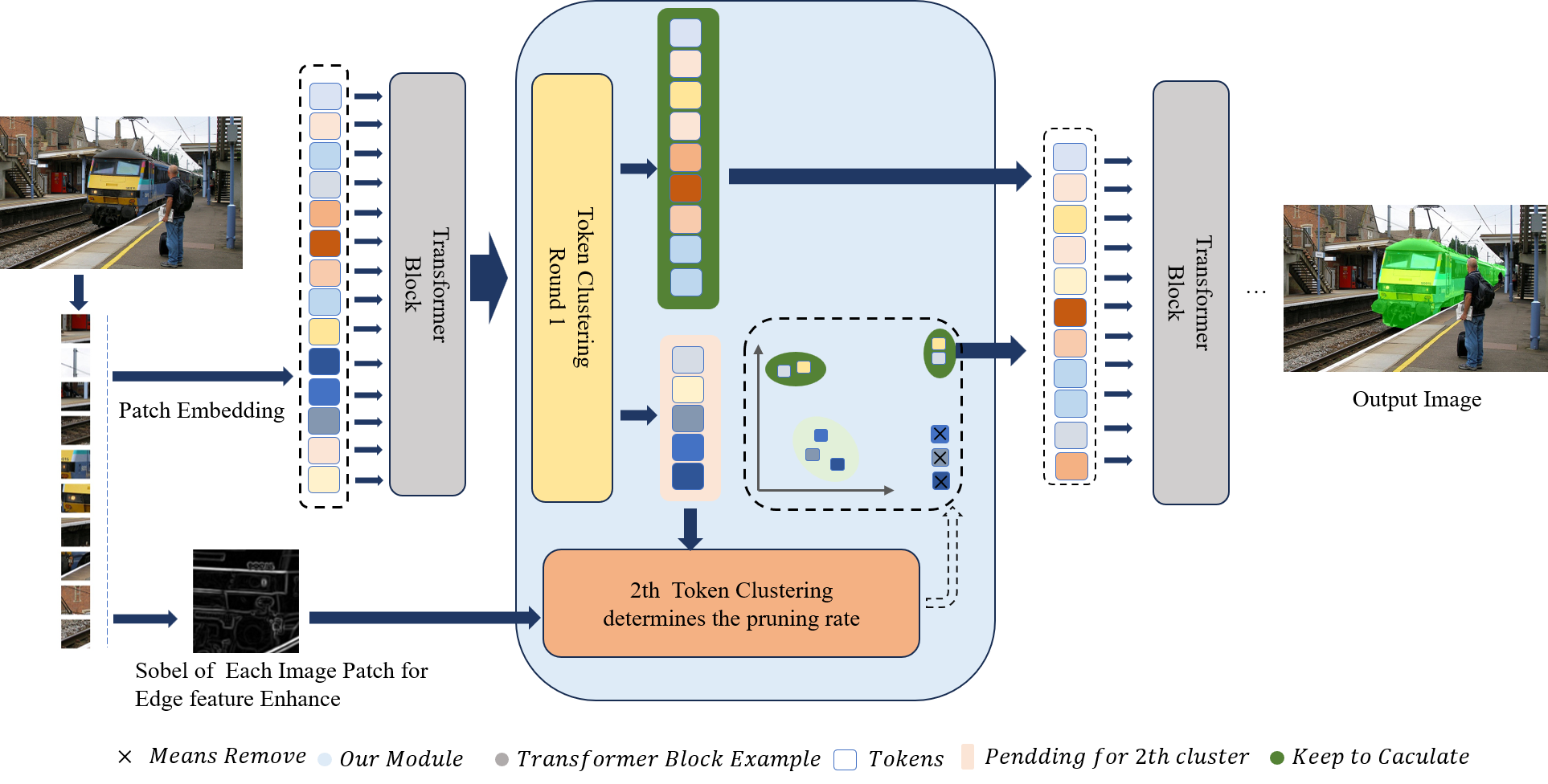} 
    \caption{Framework for Edge-Enhanced Token Clustering in Transformers}
    \caption*{Figure note: The process begins with patch embedding, where the image is divided into patches and processed by a Transformer block. After initial token clustering, Sobel edge detection refines edge features for a second clustering round, determining the pruning rate while preserving key edge tokens. This continues through Transformer blocks, progressively marking tokens for removal (×) or retention (green), enhancing segmentation inference.}
    \label{fig:framework}
\end{figure*}


\section{Methodology}

As depicted in Figure 1, the end-to-end adaptive token pruning strategy proposed in this study dynamically perceives the semantic feature distribution through multiscale Tsallis entropy. Combined with a weight adjustment mechanism that constructs an entropy-weighted mapping mechanism driven by feature energy to guide the initial coarse classification of tokens. To further retain more contextual information, we integrate the Sobel operator to enhance edge features of low-level image features, directing the final pruning decision. This approach provides an efficient and interpretable lightweight solution for ViT-based model in semantic segmentation tasks.
\subsection{ Information Entropy and Shannon Entropy}

The concept of information entropy was first introduced by Claude Shannon in 1948 to quantify the uncertainty of information. Shannon entropy is defined as:
\begin{equation}
H(X) = -\sum_{i=1}^{n} p(x_i) \log p(x_i), \label{eq:shannon_entropy}
\end{equation}

\noindent
where \( p(x_i) \) represents the probability of the \( i \)-th value of the random variable \( X \), and \( n \) is the total number of possible values. Shannon entropy reflects the average uncertainty of information; the higher its value, the greater the uncertainty of the information, and with the possibility of more semantic information.

Despite its significant success in many fields, Shannon entropy assumes that subsystems within a system are independent of each other. However, in practical applications, many subsystems interact and are correlated, such as image features. To overcome this limitation, researchers have proposed several generalized definitions of entropy, the most notable of which are R\'{e}nyi entropy and Tsallis entropy.

\subsection{Multi-scale Tsallis Entropy}

To overcome the limitation in Shannon entropy, Tsallis entropy is proposed by Constantino Tsallis in 1988\cite{39}, offers a non-extensive entropy designed to describe systems with long-range interactions or non-equilibrium states. It is defined as:

\begin{equation}
S_q(X) = \frac{1}{q-1}\left(1 - \sum_{i=1}^{n} p(x_i)^q\right),
\end{equation}

\noindent
where \( q \) is the non-extensivity parameter, which controls the correlation within the system. When \( q \to 1 \), Tsallis entropy reduces to Shannon entropy. In ViT, Tsallis entropy effectively quantifies the semantic richness of each token by capturing pixel correlations more comprehensively than Shannon entropy\cite{42}. However, traditional Tsallis entropy exhibits limitations including single-parameter rigidity that restricts multi-scale feature analysis, problematic probability sensitivity that overemphasizes high-probability regions while diminishing low-probability contributions, and insufficient modeling capacity for cross-scale feature interactions that are crucial for complex visual understanding tasks in intensive vision tasks.

To address the challenge, we proposed multi-scale Tsallis entropy, which is designed to adapt to the multi-granularity modeling needs of feature spaces. Compared to the linear assumption of feature importance in traditional single-parameter entropy systems, the multi-scale extension introduces two independent non-extensive parameters: the feature saliency factor \( q_1 \) and the detail sensitivity factor \( q_2 \). These parameters respectively adjust the sensitivity to high-probability and low-probability events to meet the needs of multi-scale analysis of complex feature distributions. The calculation form can be decoupled as follows:

\begin{equation}
S_{q_1,q_2}(P) = \alpha \cdot \frac{1 - \sum p_i^{q_1}}{q_1 - 1} + \beta \cdot \frac{1 - \sum p_i^{q_2}}{q_2 - 1}, \label{eq:multi_scale_tsallis}
\end{equation}

\noindent
where \( q_1 > 1 \) to enhance the main features with high probability and increase the response intensity to key tokens; \( q_2 < 1 \) to increase attention to low-probability regions and reduce the risk of ignoring edge detail information. \( \alpha \) and \( \beta \) are dynamically adjustable parameters to balance the two types of features.

To further avoid the limitations of rigid parameters in real-world applications, we proposed a dynamic weight adjustment mechanism. By evaluating high-norm features, \( \alpha \) is assigned as follows:

\begin{equation}
\alpha = \frac{\| F \|_2}{\sqrt{D}}, \quad \beta = 1 - \alpha
\end{equation}

\noindent
where \( F \) is the feature vector matrix of the current token, \( \| F \|_2 \) quantifies the feature activation strength of the current layer to suppress over-smoothing, and \( D \) is the feature dimension, which normalizes the high-dimensional space to avoid the curse of dimensionality. This method balances the global token information and performs comprehensive evaluation.

\subsection{Token Clustering Based on Multi-scale Tsallis Entropy}

Based on the significant redundancy of information in the image feature space, we proposed a token clustering mechanism guided by multi-scale Tsallis entropy to achieve semantic focusing in the feature space. To address the clustering bias of traditional k-means\cite{39} algorithms in high-dimensional feature spaces due to ignoring semantic importance differences, we construct a multi-scale Tsallis entropy weight matrix for feature vectors. This weighted mechanism works in three key dimensions: first, calculating the Tsallis entropy values of feature vectors at two scales; second, enhancing cross-scale semantic consistency features through entropy fusion; and finally, the normalized weights are calculated as:
\begin{equation}
\tilde{W}_{\tau} = \frac{W_{\tau} - \min(W_{\tau})}{\max(W_{\tau}) - \min(W_{\tau})},
\end{equation}

\noindent
where \(W_{\tau}\) is the entropy-guided modulation weight matrix. The enhanced feature is generated as:

\begin{equation}
\tilde{v}_i = v_i \odot \tilde{W}_\tau. \label{eq:enhanced_feature}
\end{equation}

The improved k-means algorithm uses cosine similarity as the distance metric and dynamically updates the cluster centers during iterations. By entropy-weight modulation, the algorithm generates a strong attraction to semantic focusing regions and a weak response to low-entropy redundant regions, ultimately achieving hierarchical separability in the feature space. The distance between a token \( v_i \) and a cluster center \( c_j \) is defined as:

\begin{equation}
d(v_i, c_j) = 1 - \frac{v_i \cdot c_j}{\| v_i \| \| c_j \|}, \label{eq:cosine_similarity}
\end{equation}

\noindent
where the cluster center \( c_j \) is updated as:

\begin{equation}
c_j = \frac{1}{|C_j|} \sum_{v_i \in C_j} \tilde{v}_i. \label{eq:cluster_center_update}
\end{equation}


\subsection{Low-level Visual Features Enhanced}

In semantic segmentation tasks using Transformers, preserving edge information is crucial for accurate pixel-level understanding. While shallow layers often focus on edges, not all edges of segmentation targets receive adequate attention, with some emerging only in deeper layers and potentially being eliminated during initial clustering. Additionally, backgrounds frequently share similar colors and textures with segmentation targets, making them difficult to distinguish based solely on high-level features.

To address the aforementioned challenges, Sobel\cite{43} edge detection is introduced with the objective of enhancing and guiding secondary clustering. The Sobel operator, a discrete differential operator, calculates the approximate gradient of image intensity, thereby providing critical low-level edge information that complements the high-level semantic features from Transformers. Semantic segmentation's pixel-level precision requirements make this edge-preserving approach particularly important for maintaining boundary accuracy.



The Sobel operator uses convolution kernels in the horizontal (\( x \)) and vertical (\( y \)) directions:

\begin{equation}
G_x = \begin{bmatrix}
-1 & 0 & 1 \\
-2 & 0 & 2 \\
-1 & 0 & 1
\end{bmatrix}, \quad
G_y = \begin{bmatrix}
-1 & -2 & -1 \\
0 & 0 & 0 \\
1 & 2 & 1
\end{bmatrix}
\end{equation}

By convolving the kernels \( G_x \) and \( G_y \) with the image \( I(x,y) \), we obtain gradient components that highlight edge regions where intensity changes sharply. The gradient magnitude is calculated as:
\begin{equation}
G(x,y) = G_x(x,y)^2 + G_y(x,y)^2
\end{equation}



This edge information explicitly guides secondary token clustering, ensuring that tokens containing significant edge information are preserved even if they were initially pruned. By explicitly incorporating edge features into the decision-making process,  LVTP achieves more precise boundary delineation. It improves segmentation accuracy, particularly for fine details and small objects that might otherwise be overlooked.

\begin{figure}[htbp]
    \centering
    \includegraphics[width=\textwidth]{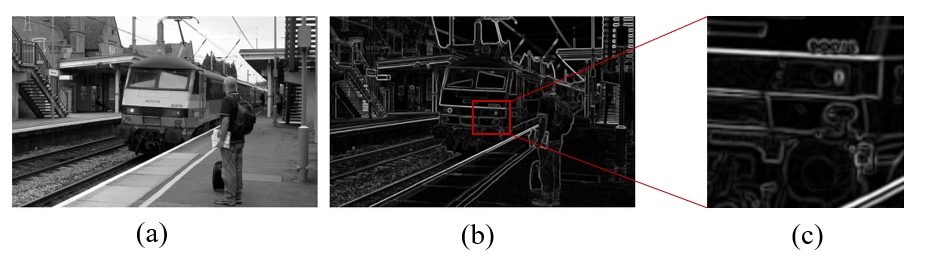} 
    \caption{Edge Enhancement Example}
    \label{fig:edge_enhancement}
    \captionsetup{font=footnotesize}
    \caption*{Figure note: Image (a) shows the original grayscale image. Image (b) is the result of applying the Sobel operator to the original image, highlighting edges and making them more prominent. The red box in image (b) indicates a region of interest where edge enhancement is particularly noticeable. Image (c) is a zoomed-in view of the region within the red box, showing the detailed edge structure after enhancement. This visual demonstration underscores the utility of edge enhancement in improving the detection and segmentation of features within images.}
    \label{fig:sobel}
\end{figure}

As illustrated in Figure \ref{fig:sobel}, the Sobel operator transforms subtle intensity changes into prominent edge features that will guide the token preservation process.

\section{Experiments}

\subsection{Datasets and Metrics}
To verify the generalization capability of the LVTP across different tasks, we selected three cross-domain datasets for testing, as detailed below:

\begin{itemize}
    \item \textbf{Massachusetts Road Remote Sensing Dataset}\cite{36}: We used to evaluate the multi-scale semantic segmentation capability in complex geographical scenes. The original images have a resolution of \(1500 \times 1500\) pixels, with a ground resolution of 1.2 meters per pixel, making it suitable for high-precision geographical scene analysis. 
    \item \textbf{COCO-Tasks Dataset}\cite{34}: This dataset contains 14 different task objectives and is used to test the adaptability of the model in multi-class dense target scenarios.
    \item \textbf{RIO Dataset}\cite{35}: This dataset is used to evaluate the effectiveness of 3D spatial feature extraction and contains over 100 diverse tasks.
\end{itemize}

These datasets cover tasks in remote sensing, general object detection, and indoor understanding, providing a comprehensive foundation for evaluating the generalization capability of LVTP across diverse domains.


To assess the robustness and efficiency of LVTP, we used a comprehensive evaluation framework with two complementary sets of metrics.

For performance measurement, we selected mean Intersection over Union (mIoU) as our primary metric for semantic segmentation quality. This metric calculates the ratio between the intersection and union of predicted segmentation masks and ground-truth regions, effectively quantifying pixel-level classification accuracy. Higher mIoU values indicate more precise segmentation results.

For computational cost, we measured the number of float-point operations in giga (GFLOPS), which represents the computational cost required by the model. Lower GFLOPS values indicate reduced computational complexity and improved inference efficiency.

To evaluate the balance between model performance  and computational cost, we defined a metric \(\gamma\) as:
\begin{equation}
\gamma = \frac{\Delta \text{GFLOPS}}{\Delta \text{mIoU}}.
\end{equation}

Where $\Delta$GFLOPS represents the reduction ratio in computational cost and $\Delta$mIoU represents the percentage decrease in segmentation accuracy. This ratio $\gamma$ quantifies the efficiency of LVTP, higher $\gamma$ values indicate more favourable trade-offs where substantial computational savings are achieved with minimal performance degradation.

Through the comprehensive evaluation of these metrics, we can fully validate the effectiveness and robustness of the LVTP across different tasks.

All of our experiments were conducted on the same machine, which contained one Nvidia RTX 4090 GPU.

\subsection{Implementation Details}
To examine the adaptability of LVTP, we conducted experiments with ViT models based on the Transformer architecture and Swin Transformer\cite{37} utilizing a hierarchical window attention mechanism as the backbone. To demonstrate the applicability of  LVTP to visual LLMs, we tested SAM\cite{38} with different backbone networks across several datasets, ensuring that no modifications were made to the models during the testing phase to ensure the objectivity of the results. The results of the benchmark performance test were displayed in the subsequent table \ref{table:generalization}. In subsequent experiments, ViT-H denotes ViT huge, ViT-L denotes ViT large, Swin-L denotes Swin Transformer large, and similarly, Swin-B denotes Swin Transformer base.

\subsection{Main Results}
\subsubsection{Compare with SOTA Token Pruning Methods}

We evaluated the LVTP against the state-of-the-art token pruning methods: CTS\cite{31}, DToP\cite{32}, SViT\cite{33}, and VLTP\cite{30} to explore the performance of each method in terms of computational cost and accuracy. All methods were not trained or fine-tuned.

\begin{table}[htbp]
\centering
\caption{Comparison of the state-of-the-art token pruning methods: Consistent comparison parameters. All experiments are based on the SAM VIT-H model with pruning operations conducted at the 16th layer.}
\begin{tabular}{lccc}
\hline
\textbf{Method} &  \(\Delta\) \textbf{GFLOPS\%} $\downarrow$&  \(\Delta\) \textbf{mIoU\%} $\uparrow$& \textbf{\(\gamma\)} $\uparrow$\\
\hline
CTS\cite{31}  & -36.4 & -24.7 & 1.47 \\
DToP\cite{32}  & -33.4 & -24.1 & 1.39 \\
SViT\cite{33}  & -35.0 & -26.9 & 1.30 \\
VLTP\cite{30} & -40.1 & -8.3 & 4.83 \\
\textbf{LVTP (Ours)} & \textbf{-46.2}& \textbf{-5.0} & \textbf{9.24} \\
\hline
\end{tabular}
\label{table: Token_compare}
\end{table}

As shown in Table \ref{table: Token_compare}, LVTP achieves a 46\% reduction in GFLOPS, outperforming all other comparison methods and surpassing the second-highest VLTP by 6.1\% in terms of GFLOPS reduction. Along with the remarkable reduction in computational cost, it exhibits the least mIoU performance degradation, with the mIoU decreasing by only 5.0\% without fine-tuning, which is significantly lower than all other methods.

In terms of the $\gamma$ metric, which represents the trade-off between computation reduction and mIoU degradation, the $\gamma$ of LVTP is 9.24, significantly surpassing all other comparison methods. This fully demonstrates the superiority of LVTP. The proposed multi-scale Tsallis entropy and low-level image feature-guided token pruning can effectively identify redundant tokens and retain tokens containing key semantic information. This allows for the retention of relatively high performance while minimizing computation as much as possible, achieving the best balance between performance degradation and computation reduction.

\subsubsection{Generalization Experiment on Different ViT Backbones and Segmentors}

\begin{table}[htbp]
\centering
\caption{Generalization Experiment on Different ViT Backbones and Segmentors: * denotes post-pruning results. For GFLOPS, the percentage represents the change rate, while for mIoU, it represents the change value. RIO, COCO, and Ma represent three datasets. Pruning is at ViT-H’s 16th layer, ViT-L’s 12th layer, and Swin-L/Swin-B’s 2nd layer.}
\begin{tabular}{lcccc}
\toprule
\small
\textbf{Model} & \textbf{Backbone} & \textbf{mIoU} & \textbf{GFLOPS}  \\
\midrule
SAM-RIO & ViT-H & 0.53 & 820   \\
SAM-RIO* & ViT-H & 0.48\,(-5\%) & 441\,(-46.22\%) \\
\midrule
SAM-RIO & ViT-L  & 0.52 & 397   \\
SAM-RIO* & ViT-L & 0.46\,(-6\%) & 230\,(-42.10\%) \\
\midrule
SAM-COCO & ViT-H & 0.32 & 820   \\
SAM-COCO* & ViT-H  & 0.29\,(-3\%) & 447\,(-45.49\%) \\
\midrule
SAM-COCO & ViT-L & 0.31 & 397  \\
SAM-COCO* & ViT-L & 0.27\,(-4\%) & 231\,(-41.8\%)\\
\midrule
Swin-Unet-Ma & Swin-L & 0.52 & 68 \\
Swin-Unet-Ma* & Swin-L & 0.52\,($\geq$-0.1\%) & 54\,(-20.59\%) \\
\midrule
Swin-Unet-Ma & Swin-B & 0.52 & 31   \\
Swin-Unet-Ma* & Swin-B & 0.51\,(-1\%) & 15\,(-19.35\%) \\
\bottomrule
\label{table:generalization}
\end{tabular}
\end{table}

To verify the generalization of the LVTP for ViT backbones of different sizes and Segmentors, we selected two transformer models: ViT and Swin-ViT, which incorporates a shifted window mechanism and has a lower computational cost. We also selected two segmentors: SAM based on ViT and Swin-Unet based on Swin-ViT, and conducted experiments on three datasets to test the pruning effect of the LVTP.

As shown in Table \ref{table:generalization}, on the RIO dataset, for ViT-H based SAM which has more layers and a complex structure, pruning was performed at the 16th layer, resulting in a significant reduction in GFLOPS (-46.22\%) with only a 5\% decrease in mIoU. This indicates that the LVTP achieves a significant reduction in computational cost and minimal performance degradation. On the COCO dataset, LVTP demonstrated similar performance, with a 45.49\% reduction in GFLOPS and only a 3\% decrease in mIoU. After pruning at the 12th layer of ViT-L based SAM, there was also a significant reduction in GFLOPS and an acceptable decrease in mIoU. This shows that LVTP can achieve a great balance between computational cost reduction and performance degradation on different sizes of ViT-based segmentors.

For Swin-Unet based on Swin-ViT, on the Massachusetts-Road dataset, after pruning the second layer of Swin-L and Swin-B, there was a 20\% reduction in GFLOPS with almost no decrease in mIoU. This indicates that the LVTP can still achieve effective pruning on Swin-ViT, which demands less computation, effectively reducing the computational cost while maintaining performance.

In summary, the LVTP can be applied to segmentors based on ViTs and their variants of different sizes to achieve effective token pruning of the model, significantly reducing the computational cost without causing a significant drop in performance. Generalization experiments have demonstrated the potential of LVTP  as an efficient plug-and-play pruning method for ViT-based models.

\subsubsection{Visualization Analysis}

\begin{figure*}
    \centering
    \includegraphics[width=\textwidth]{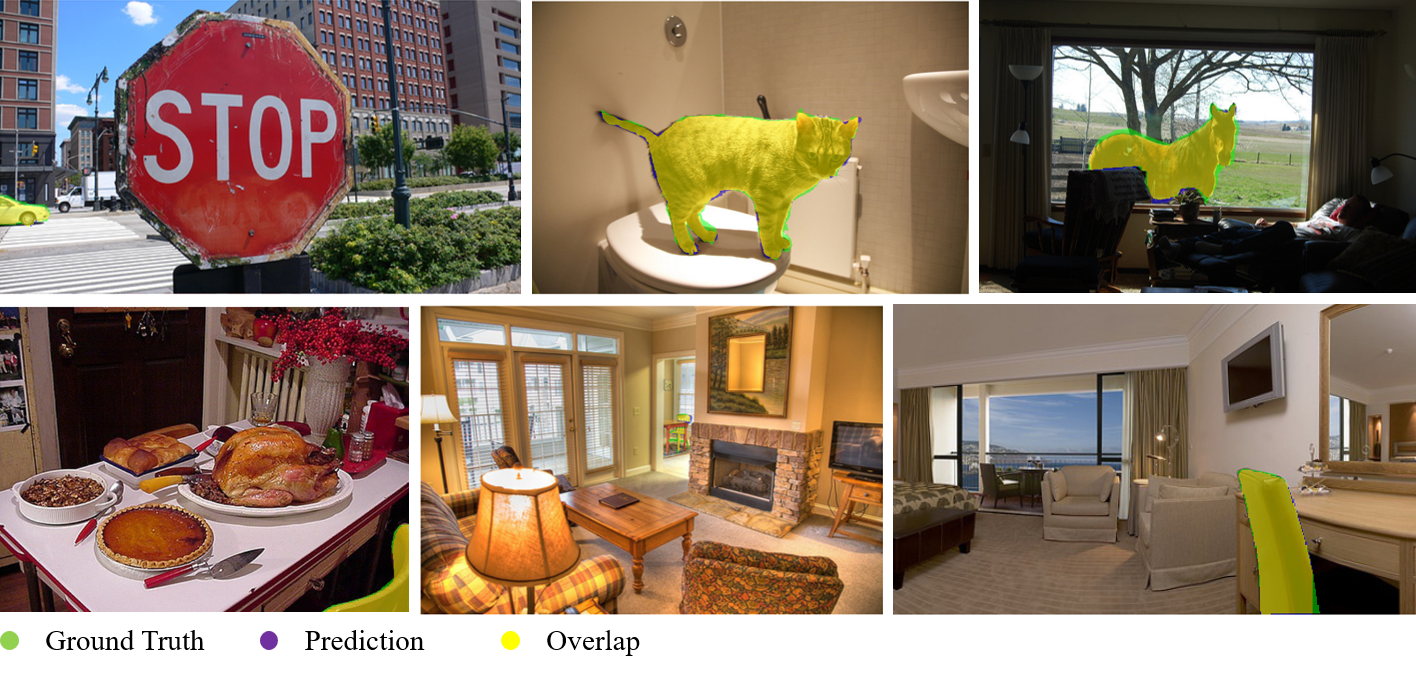} 
    \caption{Pruned SAM-ViT-H's prediction from RIO and COCO dataset}
    \label{fig:rio}
    \captionsetup{font=footnotesize}
    \caption*{Figure note: The top three images are from the RIO, and the bottom three images are from the COCO. We overlap the mask and pruned SAM-ViT-H's prediction for evaluation. Even the smaller objects (the car in the background, the chair in the distant room) achieve precise segmentation results.}
\end{figure*}

\begin{figure*}[h!]
    \centering
    \includegraphics[width=\textwidth]{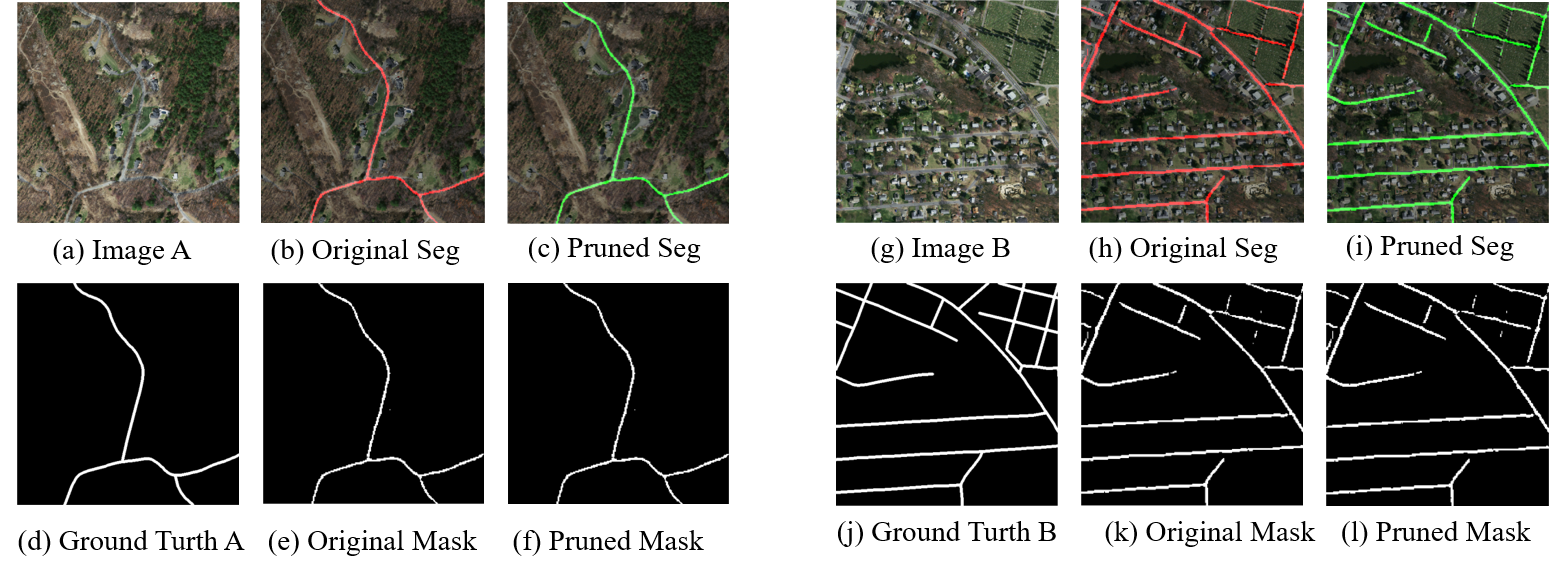} 
    \caption{Pruned Swin-Unet-L's prediction from Massachusetts-Road dataset}
    \label{fig:road}
    \captionsetup{font=footnotesize}
    \caption*{Figure note: The figure shows two sets of segmentation results, respectively, with the original image and the segmentation result superimposed on the top, from the original model and the pruned model. The bottom figures are the corresponding segmentation extractions with the above.}
    \label{fig: Mroad}
\end{figure*}

\textbf{RIO and COCO dataset, preservation of small targets}: 
Figure \ref{fig:rio} shows the overlaid images of the pruned SAM-ViT-H's predicted results and the ground truth (GT) on the RIO and COCO datasets. It can be observed that precise segmentation is achieved for smaller objects in the images, such as background vehicles and chairs in distant rooms, as well as for larger objects like cats and horses that occupy half of the image. This indicates that the LVTP, by introducing secondary clustering based on shallow texture information, retains the high-resolution details needed for recognizing small targets, enabling the model to robustly identify multi-scale targets in complex scenes even with reduced computational cost.

\textbf{Massachusetts-Road dataset, preservation of dense targets}: 
Figure \ref{fig: Mroad} compared the predicted results of Swin-Unet-L before and after pruning on the Massachusetts-Road dataset. It can be observed that for both sparse and dense road scenes, the pruned model maintains excellent segmentation performance. The pruned model precisely delineates road contours in sparse areas and effectively distinguishes complex and intertwined road networks in dense regions. This demonstrates that the LVTP does not simply prune a large number of tokens in dense prediction scenarios, but instead selectively removes irrelevant features and retains effective ones, achieving high compatibility with dense prediction tasks.

\subsection{Ablation Study}


In this section, we examined the contribution of different entropy guidance, secondary clustering, the number of pruning layers, and the pruning rate in the secondary clustering to the effectiveness of LVTP. The experimental results are presented in Table 3 to Table 6 and Figure 5.

\subsubsection {Ablation Study on Entropy Methods}
\begin{table}[h!]
\centering
\caption{Ablation study on different entropy guidance }
\begin{tabular}{lcccc}
\toprule
\textbf{Model} & \textbf{Entropy} & \textbf{mIoU(\%) $\uparrow$} & \textbf{GFLOPS $\downarrow$} & \textbf{\(\gamma\) $\uparrow$} \\
\midrule
SAM-RIO & - & 44.47 & 664  & 2.17 \\
SAM-RIO & Shannon & 42.95  & 655 & 1.95 \\
SAM-RIO &  \textbf{M-Tsallis} & \textbf{ 48.24}  &  \textbf{441} &  \textbf{9.24} \\
\midrule
SAM-COCO & - & 29.28 & 665 & 5.89 \\
SAM-COCO & Shannon & 27.89  & 655 & 4.38 \\
 SAM-COCO &  \textbf{M-Tsallis} &  \textbf{29.35} &  \textbf{447 }&  \textbf{14.68} \\
\bottomrule
\end{tabular}
\label{entropy}
\end{table}

We compared the three different entropy guidance: no entropy guidance (only tokens' feature are applied in clustering), Shannon entropy guidance, and multi-scale Tsallis entropy guidance. 
As shown in Table \ref{entropy}, we compare three entropy guidance methods: no entropy guidance, Shannon entropy, and multi-scale Tsallis entropy. Experiments are conducted on SAM ViT-H with a 50\% pruning rate applied during the second clustering. The multi-scale Tsallis entropy guidance outperformed the no entropy guidance and Shannon entropy guidance in terms of the mIoU metric, and it also leads to a greater reduction in GFLOPS, with corresponding higher $\gamma$ values. This indicated that the multi-scale Tsallis entropy guidance can effectively enhance model performance, better maintain segmentation accuracy while reducing computational cost, and achieve a balance between efficiency and accuracy.


\subsubsection {Ablation study on 2nd clustering}
\begin{table}[htbp]
\centering
\caption{Ablation study on the 2nd clustering}
\begin{tabular}{lcccc}
\hline
\textbf{Dataset} & \textbf{Method} &  \(\Delta\)\textbf{GFLOPS \% $\downarrow$} &  \(\Delta\)\textbf{mIoU \% $\uparrow$} & \textbf{\(\gamma\) $\uparrow$} \\
\hline
RIO  & \textbf{Sobel Guided clustering}  & -46.2 & -5.0& \textbf{9.24} \\
RIO  & Simple 2nd clustering & -42.5& -7.3& 5.82 \\
RIO  & NO 2nd clustering & -55.7& -11.9 & 4.69 \\
\hline
COCO &\textbf{Sobel Guided clustering}  &-45.5& -3.1 & \textbf{14.68}\\
COCO  & Simple 2nd clustering  & -41.2 & -5.1& 8.07 \\
COCO  & NO 2nd clustering & -54.6 & -9.1 & 6 \\
\hline
\end{tabular}
\label{2nd_clustering}
\end{table}

As shown in Table \ref{2nd_clustering}, "NO 2nd clustering" refers to the pruning framework performing only one clustering operation; "Simple 2nd clustering" means the same method used in the first clustering is applied for the second clustering; and "Sobel Guided clustering" refers to the second clustering being guided by low-level image features enhanced by the Sobel operator. All experimental results were obtained using the SAM ViT-H model with a 50\% pruning rate for the second clustering.

Using only primary clustering significantly reduced GFLOPS but also caused a notable performance loss, with mIoU dropping by 11.9\%, which was unacceptable. Primary clustering focused solely on tokens classified as having semantic information by entropy, thereby discarding more contextual information. In comparison, the proposed progressive clustering method improved performance. Although the GFLOPS reduction rate was reduced by 13\%, the mIoU drop was much smaller, from -11.9\% to -7.3\%.


Furthermore, the Sobel - guided clustering method based on low - level features achieved the minimal impact on mIoU performance (with only a 5\% and 3.1\% decrease on the RIO and COCO datasets, respectively), while significantly reduced GFLOPS ( - 48.2\% and - 45.5\% respectively), and achieved the highest $\gamma$ performance on both datasets. Shallow features containing more texture and edge information played a vital role in effectively guiding token clustering and pruning. By retaining tokens containing more texture and edge information, the method achieved a balance between computational reduction and mIoU performance preservation to the greatest extent.

\subsubsection {Ablation Study on Pruning Rate Analysis for 2nd clustering}

We primarily focued on the impact of different pruning rates on the model's computational cost and accuracy during the second clustering. All experiments were conducted based on the SAM ViT-H model.

\begin{table}[ht]
\centering
\caption{Ablation study on pruning rate analysis for 2nd clustering}
    \begin{tabular}{lccccccc}
        \toprule
        & \textbf{Pruning Rate} & 0\% & \textbf{20\%} & \textbf{40\%} & \textbf{50\%} & \textbf{60\%} & \textbf{80\%} \\
        \midrule
        {RIO} & GFLOPS & 528 & 491 & 457 & 441 & 425 & 394 \\
        & mIoU(\%)  & 46.5 & 47.3 & 48.5 & 48.2 & 47.4 & 44.3 \\
        \midrule
        {COCO} & GFLOPS & 532 & 495 & 462 & 447 & 431 & 401 \\
        & mIoU(\%)  & 30.3 & 30 & 29.5 & 29.4 & 29.1 & 28.6 \\
        \bottomrule
    \end{tabular}

\label{tab:pruning_rate}
\end{table}

\begin{figure*}[ht]
    \centering
    \includegraphics[width=\textwidth]{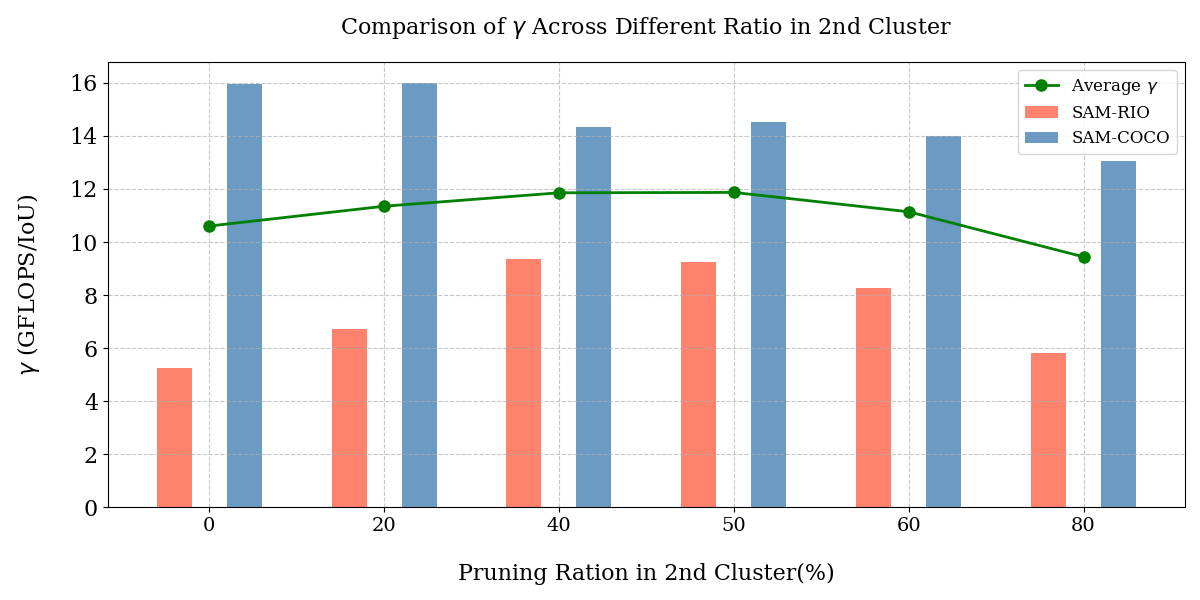} 
    \caption{Framework for Edge-Enhanced Token Clustering in Transformers}
    \label{fig:framework}
    \captionsetup{font=footnotesize}
    \caption*{Some of the important data in Table \ref{tab:pruning_rate} is visualized in the figure, which intuitively shows the average \(\gamma\) values for different pruning rates and the \(\gamma\) values in different tasks.}
\end{figure*}

As shown in Table \ref{tab:pruning_rate}, on the RIO dataset, as the pruning rate increased from 0\% to 40\%, the GFLOPS shows a gradual decreasing trend, dropping from 528 to 457, while the mIoU value shows a gradual increasing trend, rising from 46.5\% to 48.5\%. This upward trend converges at a pruning rate of 40\%. When the pruning rate was further increased, GFLOPS continued to decrease, but mIoU started to decline. The downward trend of mIoU intensified as the pruning rate increases. This indicated that in the RIO dataset, a moderate increase in the pruning rate could effectively reduce the computational consumption while improving the model's segmentation accuracy. However, an excessively high pruning rate leads to a significant decrease in accuracy.

On the COCO dataset, both mIoU and GFLOPS decreased as the pruning rate increases. When the pruning rate was low, the decrease in mIoU is relatively small, but the downward trend of mIoU becames more pronounced as the pruning rate increases. Therefore, in the COCO dataset, a suitable pruning rate helped to balance computational cost and accuracy.

Figure \ref{fig:framework} intuitively shows the average $\gamma$ of the model on the two datasets under different pruning rates. Combining the above experiment results, the advantage of a 50\% pruning rate was evident. In the RIO dataset, GFLOPS is significantly reduced while the mIoU value remained at a relatively high level. In the COCO dataset, the decline in mIoU is within an acceptable range while effectively reducing GFLOPS. This indicated that a higher pruning rate is not necessarily better, and a 50\% pruning rate could achieve the best balance between computational reduction and performance.

\subsubsection{Ablation Study on  Pruning Layers}

\begin{table}[http]
\centering
\caption{Ablation study on pruning layers: focusing on the global attention blocks of SAM ViT-H, namely 8th, 16th, and 24th layer. All experimental results are based on SAM ViT-H and involve 50\% pruning with the second clustering.}
\begin{tabular}{lcccc}
\hline
\textbf{Dataset} & \textbf{Layer} &  \(\Delta\) \textbf{GFLOPS\%} $\downarrow$ &  \(\Delta\) \textbf{mIoU\%} $\uparrow$ & \textbf{\(\gamma\)} $\uparrow$ \\
\hline
RIO  & \{8\}  & \textbf{-62.5} & -38.0 & 1.64 \\
RIO  & \{16\} & -46.2 & \textbf{-5.0} & \textbf{9.24} \\
RIO  & \{24\} & -49.7 & -25.6 & 1.94 \\
RIO  & \{8, 16\} & -59.8 & -26 & 2.3 \\
RIO  & \{16, 24\} & -52.9 & -32.7 & 1.62 \\
RIO  & \{8, 16, 24\} & -58.1 & -34.6 & 1.68 \\
\hline
COCO  & \{8\}  & \textbf{-61.8} & -23.8 & 2.60 \\
COCO  & \{16\} & -45.5 & \textbf{-3.1} & \textbf{14.68} \\
COCO  & \{24\} & -48.2 & -19.0 & 2.54 \\
COCO  & \{8, 16\} & -59.7 & -17.4 & 3.43 \\
COCO  & \{16, 24\} & -52.8 & -21.7 & 2.43 \\
COCO  & \{8, 16, 24\} & -57.9 & -22.7 & 2.55 \\
\hline
\end{tabular}
\label{layers}
\end{table}

Table \ref{layers} shows pruning effects on single/multiple layers of SAM ViT-H's global attention blocks (8th/16th/24th layer). Chosen for their role in capturing long-range sequence dependencies in Transformers, unlike local attention, global attention used the full sequence for weight computation, adding redundant info. Therefore, pruning this information was crucial for reducing the computational cost while maintaining model accuracy.

As shown in Table \ref{layers}, pruning the 16th layer achieved the best balance between reducing computational cost and maintaining model performance, delivering the optimal $\gamma$ on both the COCO and RIO datasets (9.24 and 14.7, respectively). When pruning was applied to the shallower eighth layer, GFLOPS is significantly reduced by over 60\%, but this led to a substantial decrease in mIoU (-38\% and -23.8\%), indicating that pruning too early hindered the model from extracting useful primitive features from images. Moreover, when pruning was applied to the deeper 24th layer, it disrupted the global high-level semantic features, causing difficulties in the subsequent decoding process and resulting in a significant mIoU decrease (-25.6\% and -19\%). 

Subsequent multi-layer pruning strategies also show that pruning shallow or deep layers would respectively affect the model's extraction of primitive features and disrupt high-level global semantic features, leading to unacceptable performance decrease. Therefore, extensive experiments have proven that pruning in the middle layer (the 16th layer) of SAM-ViT-H brought the best balance between GFLOPS and model performance. We believed that at the middle layer (the 16th layer), features were transitioning from low-level edge texture features to high-level semantic features. Pruning at this stage helped the model focus on the effective features of the image while extracting high-level semantic information, ignoring irrelevant features.

\section{Discussion}

This study focused on the task of ViT-based semantic segmentation, proposed a progressive token pruning method based on multi-scale Tsallis entropy and guided by low-level visual features, offering a novel optimization framework for ViT. 

We innovatively constructed a dynamic scoring mechanism based on multi-scale semantic entropy weights, transcending the limitations of traditional entropy by introducing factors of feature significance and detail sensitivity. This allowed for a precise assessment of the complexity of token semantic information, achieving more efficient semantic focusing. Additionally, we utilized the Sobel operator to enhance low-level image edge features, guiding secondary clustering—a first in integrating low-level features into token pruning decisions to prevent the loss of critical edge information. Moreover, the proposed token pruning plugin required no modification to the original model structure or additional training, offered a plug-and-play feature that greatly lowers the barrier to application.

Experimental results robustly validated the effectiveness of LVTP. Across multiple cross-domain datasets and various models,  LVTP significantly reduced computational cost; for instance, in the RIO task, the model based on ViT-H for SAM shows a 46.22\% decrease in GFLOPS after pruning, while also maintaining segmentation accuracy and even improving precision in some tasks. Compared to other methods, LVTP breaks away from an over-reliance on high-level abstract information and overcomes the drawbacks of additional training.

There are still directions for optimization. The balance of multi-scale Tsallis entropy in complex scenes for different scale features can be further refined to more accurately adapt to diverse image content. Due to the use of low-level image features, the model's generalization capability under extreme conditions, such as sudden lighting changes, severe occlusions, or high noise scenarios, needs to be strengthened. Future work could explore the integration of adversarial training and other techniques to enhance model robustness.
In terms of application prospects, LVTP shows clear advantages in inference tasks on resource-constrained terminal devices, such as remote sensing and autonomous driving semantic segmentation. It can be directly integrated into existing models to improve efficiency. Subsequent efforts could attempt to expand to other visual tasks such as instance segmentation and panoptic segmentation, and further optimize in conjunction with new hardware architectures to unlock greater application potential.

\section{Conclusions}
This study proposed a progressive token pruning method based on multi-scale Tsallis entropy and guided by low-level visual features to address the challenges associated with the application of ViT models to semantic segmentation tasks on devices with limited resources. This method innovatively incorporates a dynamic scoring mechanism and integrates low-level feature-guided clustering to achieve an end-to-end pruning framework design. Extensive experiments across various datasets and models have demonstrated that this approach effectively reduces computational cost while maintaining segmentation accuracy, outperforming other methods in terms of balance. This research contributes to the optimization of adaptive evaluation mechanisms, enhances the generalization capability in extreme scenarios, and broadens applications by integrating with new hardware architectures. It offers novel perspectives for model optimization in dense visual tasks and advances the development of the visual field within resource-constrained environments.

\section*{Acknowledgments}
This paper was supported by China Southern Power Grid Company Limited, the Jiangsu Provincial Scheme Double Initiative Plan JSS-CBS20230474 and the XJTLU RDF-21-02-008.

\section*{Author contributions}
Y. Ouyang: Development or design of methodology; creation of models. Conducting a research and investigation process, specifically performing the experiments, or data/evidence collection. Preparation, creation and/or presentation of the published work, specifically writing the initial draft (including substantive translation). Y. Liang and  Q. Li: Programming, software development; designing computer programs; implementation of the computer code and supporting algorithms; testing of existing code components. D. Wu: Management activities to annotate (produce metadata), scrub data and maintain research data (including software code, where it is necessary for interpreting the data itself) for initial use and later reuse. X. Guo: Preparation, creation and/or presentation of the published work, specifically writing the initial draft (including substantive translation). Y. Luo: Provision of study materials, reagents, materials, patients, laboratory samples,instrumentation, computing resources, or other analysis tools. H. Wang: Provision of study materials, computing resources, rewriting and editing the paper. Y. Pan: Ideas; formulation or evolution of overarching research goals and aims. Development or design of methodology; creation of models. Oversight and leadership responsibility for the research activity planning and execution, including mentorship external to the core team.  

\section*{Data Availability Statement}
The datasets analyzed during the current study are available from the corresponding author on reasonable request.

\section*{Conflict of Interest}
The authors declare that they have no conflict of interest.

\section*{Declaration of Interest Statement}
The authors declare that they have no known competing financial interests or personal relationships that could have appeared to influence the work reported in this paper.

\bibliographystyle{unsrt} 
\bibliography{main}

\end{document}